\pdfoutput=1
\documentclass[11pt]{article}

\usepackage[preprint]{acl}

\usepackage{times}
\usepackage{latexsym}
\usepackage[T1]{fontenc}
\usepackage[utf8]{inputenc}
\usepackage{microtype}
\usepackage{graphicx}
\usepackage{booktabs}
\usepackage{amssymb}
\usepackage{makecell}

\title{Boundaries Agree, Labels Do Not:\\ Intra-Annotator Dynamics as a Kind of Training Data}

\author{Marharyta Shvets \\
  Eötvös Loránd University (ELTE), Budapest, Hungary \\
  \texttt{rozovslonik@student.elte.hu}}

\begin{document}
\maketitle

\begin{abstract}
Data quality now matters as much as compute for training language models. Much training data comes from human annotation of text, and interpretive annotation has no ground truth that could settle what is ``accurate''. Two lines of work respond to this. One combines annotators into a ``ground truth'' and measures how well they agree with each other; the other treats their disagreement as a signal. Both compare different people at one point in time. We measure something else: how well one reader reproduces their own reading of the same text over time. One expert human reader and three LLM families segmented three Sumerian myths and labelled the causal function of each segment with one of seven states. Across runs months apart, the human cut the text in much the same places but named the segments differently, in every myth. The models show no such consistent pattern: their gap between the two layers is positive in some myths and negative in others, and its size varies. The human's label changes are not random: the runs go through much the same functions but start them one step apart, while model runs start them at the same places. We argue that this pattern is a usable measure of data quality and a contamination check: a ``human'' annotation whose labels are as stable as its boundaries, and whose functions start in sync, looks like a model's.
\end{abstract}

\section{Data quality, and where it is measured}

Besides compute, the quality of training data largely decides how well language models perform. Compute grows along a fairly predictable curve \citep{hoffmann2022}. Data can be obtained in two ways: by collecting text that already exists, or by producing new kinds of data. We argue for more data of the second kind --- data that records the logical structure a human reader builds: where one event ends and what causal function it has. This differs from a description of what is already on the page, such as a paraphrase or a summary. Unlike such descriptions, built structure can be checked formally, for example for whether it can be reproduced.

Human annotation of text is a common way to build training data for LLMs \citep{ouyang2022}. Usually the labels of several annotators are combined into a ``ground truth'', and quality is reported as agreement between annotators \citep{artstein2008}. A second line of work treats disagreement between annotators as a useful signal rather than noise --- the perspectivist turn \citep{aroyo2015,plank2022,ivey2025,xu2026,chen2026}. Both lines compare different people. How well an annotator agrees with themselves is measured far more rarely, although it helps to separate inconsistency from real label variation \citep{abercrombie2025}. We measure it in a task where each annotation has two layers, segment boundaries and labels, and show that the two layers behave differently.

Our claim is simple. When an expert human reader annotates the same text again months later with the same scheme, the segment boundaries mostly stay the same, but many labels change. The three LLM families repeat both. We call this split between the two layers a \emph{dissociation}. The human's label changes are not random either: the runs go through much the same functions, shifted by one step. We argue that both patterns are signals of data quality that neither existing line of work can detect.

\section{What is being measured: attributed causality, no ground truth}

We examine whether a human reader leaves a causal signature --- a personal way of wiring events --- when working with an object, here a text. To make this precise, we use a working definition of causality from computer science: the principle by which one event or variable directly produces or changes another.

The object is the text, and it stays an object. The event is not a fact in the text: it is the subject's reading and naming of the logic of a segment. We call the human a subject (and a reader) and a model only a reader. Both read, but only the human is assumed to bring a subject's own logic to the text; the metrics are built to test for its traces, not to assume them. The causal relations we measure therefore link events that a reader attributes. This is attributed causality in Hume's sense, not interventional causality in Pearl's \citep{hume1748,pearl2009}; readers of narrative likewise build causal links between the events they construct \citep{trabasso1985}. Hence there is no ground truth, and the structure is imposed, not read off. Every metric therefore compares a reader with themselves or with chance. A ``signature'' is a stable, individual way of attributing structure, and we can speak of one only if it reproduces within one reader. Self-consistency measures exactly this, in the same way for a human and for a model.

\section{Data and method}

\subsection{Corpora}

We use three Sumerian myths --- the Gudea Cylinders, Inanna's Descent, and Inanna and Enki\footnote{RIME 3/1.01.07, Cyl A composite (P431881); CDLI\_Literary\_000343\_Inanna\_composite\_P468903; ETCSL\_1.3.1\_EN \citep{etcsl}; texts from CDLI (\url{https://cdli.earth}) and ETCSL.} --- as the tool for our measurements, because of a combination of properties. The texts were written on clay c.\ 2200--1600 BCE, and the roots of their meaning may go back further. There is no living tradition of reading them: Sumerian is a language isolate with no descendants \citep{michalowski2004}, and it is understood today largely through Sumerian--Akkadian bilingual texts. Unlike Greek mythology or the Bible, these myths have few retellings or popular commentaries. We use the online versions of the English translations\footnotemark[\value{footnote}], which may be part of the models' training data. However, there is no tradition of dividing these myths into episodes that a model could copy. The corpora therefore show the logic that readers build while reading, rather than logic they extract from the meaning of the text. The translations were cleaned and given line numbers for machine analysis; these versions, the annotations and the code are available at \url{https://github.com/Malificenta883/intra-annotator-dynamics}.

\subsection{Readers and runs}

The readers are three model families and one expert human. The models were Claude Opus 4.8 (high reasoning), Gemini Pro and Qwen 3.8 Max (thinking), used through their web interfaces in anonymous or temporary chats with default settings, June--August 2026; the exact version and date of every run are recorded in its data file. Each model family ran 10 times in fresh sessions, in reasoning mode, with the same prompt. The human annotated each myth 3 times, with runs from five weeks to ten months apart, which keeps memory effects small. The prompt has two variants: text lines for Gudea and Inanna's Descent, blocks for Inanna \& Enki. The block variant also tells the model that gaps in numbering are lacunae, and it differs in minor formatting. Opus on Inanna's Descent received an earlier version with the same segment definition and states, which asked for line ranges in the segment title and one extra field; a script moved the ranges into line numbers. Model runs are thus repeated almost at once, while human runs are months apart; we treat these as two different kinds of reliability. LLM labels are known to vary across repeated runs of the same prompt \citep{reiss2023}.

\subsection{The annotation instrument (fixed schema)}

Each reader divides a text into segments and gives every segment one of seven function states --- \textit{preparation, contact, exchange, disruption, negotiation, stabilization, return} --- together with a transition from and to these same states. The states were derived from Inanna \& Enki, whose plot turns on the transfer of the ``me'' (the divine powers and institutions of civilization): transfer, or movement between states, is the central idea of this myth. Readers received the state names only.

There is deliberately no answer key. Function schemes of this kind go back to Propp's morphology \citep{propp1968}. Such schemes are built by hand by one scholar; they are very hard to reproduce independently, and computational work has treated the expert's analysis as the gold standard to recover \citep{finlayson2016}. We do the opposite: we measure how well one expert reading reproduces when the same reader repeats it over time.

Two features make the label metrics meaningful. First, all runs and all readers use the same seven states. So when a reader's labels change between runs, the reader has re-interpreted the same categories; the categories themselves did not change. Second, readers saw only the names, so the labels show how each reader fills a bare category name. The human reader is also the author of the scheme, so their unstable labels cannot be explained by unfamiliarity with the categories.

\subsection{Projection onto shared units}

Readers cut texts into segments of different sizes. To compare them, we map every reading onto the same small units: each unit takes the function of the segment that contains it. For Gudea and Inanna's Descent the unit is a text line. For Inanna \& Enki it is a block: the translation is organised in 71 blocks aligned with the Sumerian original. Segments follow block edges; in the one case where an edge falls inside a block (in one human run), the block takes the label of the segment that covers most of it. Lacunae, manuscript variants and lines that contain only an ellipsis are not units. A unit is a boundary if a segment begins in it. Two runs are compared on the units that both have annotated. This mapping does not hide differences in segment size; they remain visible in the boundary layer.

\subsection{Metrics (chance-corrected)}

\paragraph{Self-consistency.} We measure how well a reader agrees with their own earlier runs. For labels, we compare two runs unit by unit and compute Cohen's $\kappa$ over the seven states \citep{cohen1960}. For boundaries, we do the same with two classes: a unit either starts a segment or does not. $\kappa$ corrects for agreement expected by chance; this matters for boundaries, because most units are not boundaries in either run. Using $\kappa$ for both layers puts them on one scale. $\kappa$ gives no credit when two runs place a boundary one unit apart, while segmentation measures do \citep{lamprier2007,fournier2013}. Our boundary scores are therefore, if anything, too low, so the gap between boundaries and labels is not produced by the metric.

\paragraph{Label concentration (B1).} Do the runs disagree on the same few units? For each unit we count the runs that depart from its most common label. We compare the number of units where all three runs disagree with a baseline in which the same number of departures is scattered at random over the text, in stretches of the observed length ($N = 1000$). The baseline keeps the amount of disagreement fixed and changes only where it falls.

\paragraph{Neighbour-licensed flips (B2).} When one run gives a passage a different label, is that label taken from the passage before or after? We use only units where at least two runs agree (the majority label). Neighbouring units with the same majority label form a passage. Each time a run relabels a passage counts as one event, however long the passage. The baseline draws the new label at random from the other states ($N = 1000$).

\paragraph{Phase shift.} Do two runs start the same function at the same place? We cut the text at every boundary of every run of one reader and call the resulting stretches \emph{pieces}. Whenever a run switches to a new function, we find where another run starts the same function and count the distance in pieces: 0 (same place), 1, 2--3, or no match. For the baseline we shift one run along the text by a random number of pieces ($N = 1000$), which keeps its sequence but breaks the alignment. For the models we take every set of three runs out of ten (120 sets, 200 draws each), so that pieces are cut by three runs, as for the human. This analysis is exploratory: we noticed the pattern while inspecting the human's runs and designed the test afterwards.

\paragraph{Transferable grammar.} Does a reader follow the same order of functions in every myth? We learn how often each function is followed by each other function in two myths and test how well this predicts the order in the third. We report the gain in bits per transition over simply knowing how frequent each function is. The baseline shuffles the order of segments in the test myth ($N = 1000$). This is the only metric that compares myths; all others work within one text.

\section{Results}

\begin{figure*}[t]
  \centering
  \includegraphics[width=\textwidth]{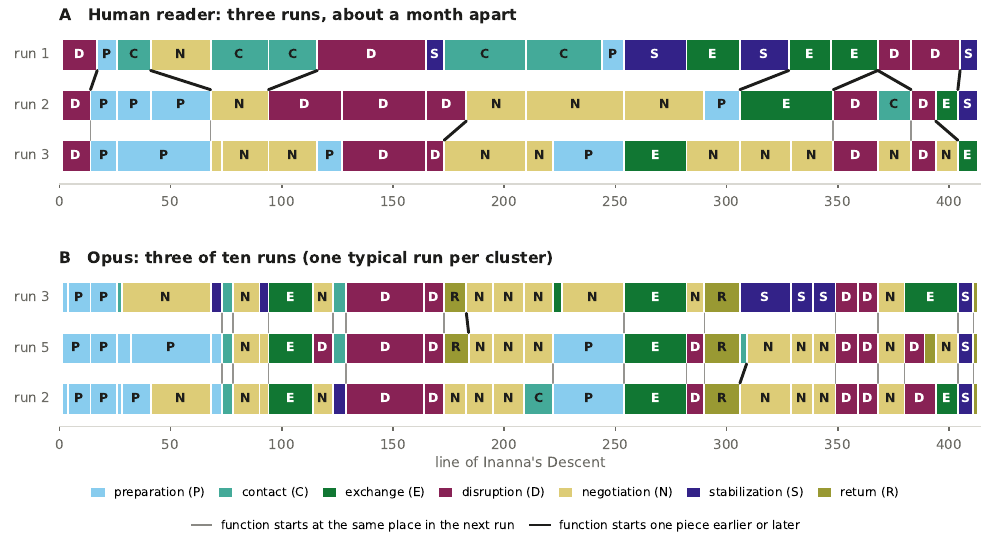}
  \caption{When does a function begin? Each bar is one run of Inanna's Descent, coloured by function; gaps are segment boundaries. Lines connect the places where neighbouring runs begin the same function: grey, at the same place; black, one piece earlier or later. (A) The human's runs go through much the same functions, shifted by one step. (B) A typical set of three Opus runs starts its functions at the same places.}
  \label{fig:phase}
\end{figure*}

\subsection{The dissociation is a human signature}

In the human reader, boundaries reproduce much better than labels in every myth (Table~\ref{tab:human}). The gap between boundary $\kappa$ and label $\kappa$ is +0.440 (Inanna and Enki), +0.452 (Inanna's Descent) and +0.298 (Gudea). Between 47\% and 78\% of one run's boundaries fall exactly on a boundary of another run; chance would give 3--21\%. People agree with one another on where events begin \citep{newtson1973,zacks2007}, and the same person places boundaries in similar places over more than a year \citep{speer2003}. What is new is that the labels on these boundaries are not stable: the human's label $\kappa$ is 0.17--0.29, against 0.50--0.85 for the models.

\begin{table}[t]
  \centering\small
  \begin{tabular}{lccc}
    \toprule
    corpus & boundary $\kappa$ & label $\kappa$ & gap \\
    \midrule
    Inanna \& Enki   & 0.731 & 0.291 & +0.440 \\
    Inanna's Descent & 0.655 & 0.203 & +0.452 \\
    Gudea            & 0.465 & 0.168 & +0.298 \\
    \bottomrule
  \end{tabular}
  \caption{Within-human self-consistency by corpus (chance-corrected $\kappa$).}
  \label{tab:human}
\end{table}

The models show no such consistent gap (Table~\ref{tab:gap}): it is positive in some myths and negative in others. Opus and Gemini are negative on Inanna and Enki, where their labels reproduce better than their boundaries. Gemini's gap on Gudea (+0.291) is as large as the human's. What sets the human apart is that the gap is large and positive in all three myths.

\begin{table}[t]
  \centering\small
  \setlength{\tabcolsep}{4.5pt}
  \begin{tabular}{lccc}
    \toprule
    reader & Inanna \& Enki & Inanna's Descent & Gudea \\
    \midrule
    human  & +0.440 & +0.452 & +0.298 \\
    Opus   & $-$0.095 & +0.160 & +0.065 \\
    Gemini & $-$0.178 & +0.032 & +0.291 \\
    Qwen   & +0.084 & +0.106 & +0.077 \\
    \bottomrule
  \end{tabular}
  \caption{Boundary--label gap by reader (within-source), per corpus.}
  \label{tab:gap}
\end{table}

\subsection{Why the time between runs does not explain it}

The human's scores are lower than the models' on both layers. This difference in level is partly due to timing: human runs are months apart, model runs almost simultaneous. The dissociation, however, compares two layers within the same human runs, measured at the same spacing, so timing cannot produce it. We therefore compare the level of human and model scores only qualitatively.

\subsection{How the human's labels disagree: a shift in phase, not a map of hard spots}
\label{sec:phase}

Does the human's label disagreement sit on a few hard units? It does not. With the total amount of disagreement fixed, units where all three runs disagree are rarer than under random scatter (ratio 0.85, 0.88 and 0.72; Table~\ref{tab:b1b2}). A baseline that shuffles labels across units gives ratios above 1 (1.90, 1.57, 1.27). But shuffling also increases the amount of disagreement, so these ratios only repeat that the labels agree more than chance.

B2 asks which label a run chooses when it relabels a passage. For the human, it is the label of the neighbouring passage in 60\% of events in Gudea (chance: 25\%, $p = 0.003$); the other two myths have too few events to decide (12 and 9). All three model families do the same, often more strongly. We therefore see B2 as a property of the scheme --- neighbouring stages of a conflict are easy to confuse --- and not as part of the human signature.

\begin{table}[t]
  \centering\small
  \setlength{\tabcolsep}{4.5pt}
  \begin{tabular}{lcccc}
    \toprule
    corpus & \multicolumn{2}{c}{B1 (fixed volume)} & \multicolumn{2}{c}{B2} \\
           & ratio & $p$ (above) & gap & $p$ \\
    \midrule
    Inanna \& Enki   & 0.85 & 0.85 & +0.149 & 0.268 \\
    Inanna's Descent & 0.88 & 0.88 & +0.159 & 0.194 \\
    Gudea            & 0.72 & 1.00 & +0.346 & 0.003 \\
    \bottomrule
  \end{tabular}
  \caption{Human label disagreement. B1: concentration against a baseline with the same amount of disagreement (below 1: less concentrated than random scatter). B2: neighbour-licensed flips.}
  \label{tab:b1b2}
\end{table}

The disagreement is organised in time. When the human starts a function in one run, another run rarely starts it at the same place: in Inanna's Descent and Gudea, starts at the same place are at chance level (8 vs.\ 6.3 expected; 14 vs.\ 8.9). Instead, the other run starts the same function one piece earlier or later, 2.5 and 1.8 times more often than chance (30 vs.\ 12.0; 32 vs.\ 17.9; $p = 0.001$). Lags of two or three pieces are at or below chance (Table~\ref{tab:phase}, Figure~\ref{fig:phase}). The runs go through much the same functions, but out of step by one piece. The models differ. In every set of three model runs, starts at the same place are well above chance (at least 2.1 times the expected count). No model triple is as out of step as the human, and on Descent and Gudea none has as many one-step lags per same-place start. Models also show one-step lags, but on top of many starts at the same place. On Inanna \& Enki, scored on blocks, the human's starts also fall at the same places (20 vs.\ 7.7). There the disagreement takes another form: in 8 of 15 unstable passages, two runs use the same two functions in opposite order on neighbouring blocks. The weak alignment partly reflects the human's low label agreement in general. What is new is its form: a shift by one step, not a random scatter.

\begin{table*}[t]
  \centering\small
  \begin{tabular}{lcccc}
    \toprule
    & \multicolumn{2}{c}{human} & \multicolumn{2}{c}{model triples} \\
    \cmidrule(lr){2-3}\cmidrule(lr){4-5}
    corpus & \makecell{same place\\obs / exp (ratio)} & \makecell{one step\\obs / exp} & \makecell{lowest\\same-place ratio} & \makecell{one-step per same-place:\\human / model max} \\
    \midrule
    Inanna \& Enki   & 20 / 7.7 (2.6) & 16 / 15.4 & 3.8 & 0.80 / 1.00 \\
    Inanna's Descent & 8 / 6.3 (1.3)  & 30 / 12.0 & 2.7 & 3.75 / 1.86 \\
    Gudea            & 14 / 8.9 (1.6) & 32 / 17.9 & 2.1 & 2.29 / 1.41 \\
    \bottomrule
  \end{tabular}
  \caption{Where two runs of one reader start the same function: at the same place, or one piece earlier or later, compared with a baseline that shifts one run along the text. Models: the most extreme value over all 360 sets of three runs per myth (200 draws each).}
  \label{tab:phase}
\end{table*}

\subsection{A transferable event-grammar: models carry it across myths, the human does not}
\label{sec:grammar}

This test asks whether a reader uses the same order of functions in every myth. From two myths we learn how often each function follows each other function, and we use this to predict the order in the third myth. A gain above zero means the prediction beats simply knowing how often each function occurs. Ranked by mean gain (Table~\ref{tab:grammar}), the human comes last (+0.04 bits per transition) and the models higher (Opus +0.44, Qwen +0.17, Gemini +0.13). Opus transfers to all three held-out myths ($p = .001$). The human transfers only to Gudea (+0.043, $p = .006$); on the other two myths the result is no better than a shuffled order. The human has 3 runs (45--74 transitions per held-out myth) and the models 10 (200--417), but with three model runs, chosen in either of two ways, the order stays the same: Opus $>$ Qwen $>$ Gemini $>$ human. Some gains are below zero (Gemini on Gudea at every number of runs, Qwen on Descent with some choices of runs): there, knowing two myths predicts the third worse than label frequencies alone. The permutation test does not check whether a gain is below zero, so we do not call these gains significantly negative.

\begin{table}[t]
  \centering\footnotesize
  \setlength{\tabcolsep}{2.8pt}
  \begin{tabular}{lcccc}
    \toprule
    reader & Gudea & I.\ \& Enki & Descent & mean \\
    \midrule
    Opus   & +0.441 {\scriptsize(.001)} & +0.522 {\scriptsize(.001)} & +0.357 {\scriptsize(.001)} & +0.440 \\
    Qwen   & +0.258 {\scriptsize(.001)} & +0.453 {\scriptsize(.001)} & $-$0.215 {\scriptsize(.001)} & +0.165 \\
    Gemini & $-$0.237 {\scriptsize(.002)} & +0.419 {\scriptsize(.001)} & +0.213 {\scriptsize(.001)} & +0.132 \\
    human  & +0.043 {\scriptsize(.006)} & +0.014 {\scriptsize(ns)} & +0.062 {\scriptsize(ns)} & +0.040 \\
    \bottomrule
  \end{tabular}
  \caption{Transferable event-grammar, one myth held out at a time. Cell: mean gain in bits per transition over label frequencies alone (permutation $p$: real order vs.\ shuffled order); higher means more transferable.}
  \label{tab:grammar}
\end{table}

In short: the models carry one event-grammar across myths, while the human's grammar stays within each myth (\S\ref{sec:phase}); we read the human as ranked lowest, not as lacking a transferable grammar.

\section{The dissociation as a data-quality axis}

The two layers answer two different data-quality questions. The segmentation layer reproduces in every reader, so it can serve as reliable structure. The labeling layer is where interpretation happens and where a human reader varies.

Combining annotators removes exactly this variation: it averages readers toward one ``ground truth'' and erases the interpretive layer. Recording one expert's own labels of the same text over time, on stable segmentation, gives data that combined annotations cannot give: not the consensus answer, but the logic by which a human reader gives a causal function to a segment, including places where more than one reading is legitimate.

This also gives a sharper quality check than a single threshold. The human signature is not low self-consistency but the gap between stable boundaries and unstable labels. A ``human'' annotation whose labels are as stable as its boundaries looks like a model. The phase shift (\S\ref{sec:phase}) suggests a second, finer check: a human's repeated labels tend to start the same functions one step apart, while model runs start them at the same places. Agreement between annotators cannot provide either check. Both matter because AI-assisted annotation is already common: 33--46\% of crowd workers in one study used LLMs for writing tasks \citep{veselovsky2023}.

\section{Discussion}

Within what the data can show, segmentation reproduces in every reader, while labeling is where interpretation --- and the difference between human and models --- appears. This agrees with the distinction between bottom-up and top-down contributions to event segmentation \citep{kurby2008}, and with the finding that people reliably notice both changes of action and changes of goal \citep{su2024}. The transfer result (\S\ref{sec:grammar}) agrees with models that carry a fixed, template-like event-grammar and a human whose grammar adapts to each text. A stronger claim --- that the models copy the organisation of reading without its intentions \citep[cf.][]{byrne1998} --- would need a panel of readers and matched sampling.

A narrower caution concerns interpretability. Stable structure is not evidence of understanding \citep[cf.\ reliability without validity in LLM judges;][]{norman2026}. A test that rewards stable, templated answers will score a model highly on exactly the layer that every reader reproduces, and it will miss the labeling layer, where the human's interpretation is variable and individual.

\section{Next steps}

A small panel of expert readers, each with at least two runs spaced in time, would test whether the features found here --- the gap between boundaries and labels, and the one-step phase shift --- appear in other people, or whether each reader has features of their own that still differ from the models'. The same data would allow fitting a generative model of reading, with separate variances for the two layers and a phase shift per reader, tested on held-out readers. With one reader and three runs, such a model would only repeat the descriptive statistics, so we do not fit it here. A re-test after a short interval would separate reinterpretation over months from ambiguity in the moment, and would make human and model reliability directly comparable.

\section*{Limitations}

The dissociation rests on one expert human reader. In effect it is replicated three times, once per myth, not once per pair of runs: within a myth the pairs share runs, so the $p$-values for pairs are too optimistic. This is a case study of a method, not a claim about people in general; a panel would be needed to tell whether the pattern is typical of experts or specific to this reader.

The human--model comparison is uneven: human runs are months apart, model runs almost simultaneous. We therefore keep it qualitative. The main claim, the dissociation within the human reader, compares two layers measured at the same spacing.

Several limits come from the metrics themselves. (a)~Comparing at equal numbers of runs is only partly fair: even the most different model runs agree with each other more (0.46--0.85) than the human's three runs do (0.32--0.38), so no model triple varies as much as the human's. (b)~Label concentration (B1) is tested with three runs only. It cannot see two runs moving a hard unit to the same label, so it could miss concentration; it finds none. (c)~B2 uses only units with a majority label; with three runs about a third of the units (30--35\%) have none and are left out. (d)~Inanna \& Enki has only 71 blocks, so every metric has little data there; for B2 this leaves 9 events, too few to decide. (e)~Boundaries reproduce least on Gudea: about half of them are placed differently across runs, not just shifted, so ``the same cuts'' holds for Gudea only in part. (f)~Permutation $p$-values rest on 1,000 draws (200 per model triple for the phase shift), so values near 0.01 are bounds, not exact values. (g)~The phase-shift counts are small (11--22 starts per run), each pair of runs is counted in both directions, and a piece can be one line or thirty lines long. We report the phase shift as a description, not as a calibrated effect.

\section*{Use of AI assistants}

Claude (Opus family; Anthropic) was used extensively to write the analysis code, run the statistical computations, and edit the text. The research question, the annotation scheme, the human annotations, the choice of metrics, and the interpretation of results are the author's; the author verified the reported results, and the code is released for inspection. The same model family is also one of the three readers studied: its annotation runs, like those of the other models, were produced in separate fresh anonymous sessions with the same prompts, and the assistant did not produce or alter any annotation.

\bibliography{references}

\end{document}